# GLR-PARSING OF WORD LATTICES USING A BEAM SEARCH METHOD

Steffen Staab
e-mail: snstaab@immd8.uni-erlangen.de
Äußere Brucker Str. 45
91052 Erlangen
Germany

## ABSTRACT

The process of understanding spoken language requires the efficient processing of ambiguities that arise by the nature of speech. This paper presents an approach that allows the efficient incremental integration of speech recognition and language understanding using Tomita's generalized LR-parsing algorithm. For this purpose the GLR-lattice-parsing-algorithm [11] is revised so that an agenda mechanism can be used to control the flow of computation of the parsing process. Subsequently the HMM-evaluations of the word models are combined with a stochastical language model to do a beam search similar to [2, 1, 12], where chartparsers are used to do the job.

## 1. INTRODUCTION

In [10] M. Tomita proposes a parsing algorithm (Generalized LR-Parsing, GLRP) and extends it in [11] to an algorithm that can parse whole word lattices. This algorithm often works more efficiently with grammars for natural languages than others (see [10, 7]).

Nevertheless the lattice-GLRP is not very flexible and requires the parse of the whole lattice in a certain order. Therefore it remains impossible to use it in real size applications that must handle 500 and more word hypotheses in each word lattice — in spite of its efficiency.

It is generally acknowledged that the problem regarding the size of the word lattices can only be solved by using heuristics that can guide the parsing process. Typically, the following two models are combined: an *acoustic model* that represents the probability that a certain word was uttered during a time interval and a *language model*, e.g. a probabilistic regular grammar that scores word sequences.

In section 2 *a revised version of the original GLRP* is presented. This new algorithm consists of three basic actions that act on the core data structure of the GLRP, the graph structured stack. They are designed in such a way that their instances may be processed in any random order, which makes it possible to put them into a control data structure (*agenda*) and work them down according to a certain strategy. This means a new quality of control over the order of processing of a parser that uses LR-parsing tables.

Also, by this way it will be possible in the sections following to combine the basic actions with a heuristic scoring function. This combination will allow to guide the search through the lattice, in order to find the word sequence that has the best evaluation and its syntactic derivation very fast.

## 2. THE REVISED GLRP

First the crucial data structures are defined to be either sets of some kind or pascal-like records:

- VERTEX *(Time, State, LinkSet)*: A *Vertex* represents a left context. It can be referenced by its time and the slr-table state it represents. Furthermore, it has got a set of links.

- LINK *(Node, PS)*: A *Link* is a reference to a node in the *ParseForest* that also connects a vertex ($A$) to a set of predecessor vertices ($PS$). The slr-table-lookup of the state of these vertices $PS$ together with the category of *Node* yields the state of vertex $A$.

- NODE *(Cat, Start, End, Hypos)* or *(Cat, Start, End, SubtrSeqs)*: It can be uniquely identified by the triple *(Cat, Start, End)*. It either references a set of word hypotheses – then the category is a terminal – or a set of sequences of subtrees – then *Cat* is nonterminal.

- GRAPH-STRUCTURED STACK, *GSS*: The *GSS* is a set of links and vertices.

- SET OF NODES, *ParseForest*: The *ParseForest* represents all possible derivations.

- SET OF ACTIONS, *Agenda*: All actions are placed on the *Agenda* and are carried out according to some strategy.

- SET OF NEWHYPOS, *OldHypoActions* that have already been executed.

The new approach does not only lead to a more flexible algorithm it also divides the work that has to be done by the GLRP more concisely into the three main mechanisms:

1. **Shift**: construct a new element in the *GSS* — if it does not already exist

2. **Search**: initiate new **Shifts** with non-terminal categories

3. **NewHypo**: initiate new **Shifts** with terminal categories

The basic actions are:

Shift(*Vertex, Node, Time, State*):

1. if $\exists V_i \in GSS$, st. $V_i = (T_i, S_i, LinkSet_i)$, $T_i = Time$ and $S_i = State$, then

    if $\exists L_j \in LinkSet_i$, st. $L_j = (N_j, PS_j)$ and $N_j = Node$, then 2. else 3.

    else 4.

2. • if $Vertex \in PS_j$ then return

    • add $Vertex$ to $PS_j$

    • assume $Vertex = (., State_l, LinkSet_l)$

    • for all previous Search(*$Rule_{nr}$, SubtrSeq, Link, EndingTime*), st. $Link = L_j$ do

        if $|SubtrSeq| = |\text{right-side}(Rule_{nr})|$ then

            add Shift(*Vertex, $N_m$, EndingTime,*
                    slr-table($State_l$, HeadCat($Rule_{nr}$)))
            to *Agenda*, where $N_m$ is the node that was created by the shift action which was initiated by this previous search

        else

            for all links $L_k = (N_k, .) \in LinkSet_l$ do

                add Search(*$Rule_{nr}$*, cons($N_k$, *SubtrSeq*), $L_k$, *EndingTime*) to *Agenda*

    • return

3. • create link $L_j = (Node, \{Vertex\})$

    • for all actions Search(*$Rule_{nr}$, SubtrSeq, Link*), st. $Link = (.,PS)$, $V_i \in PS$ and $|SubtrSeq| < |\text{right-side}(Rule_{nr})|$, do

        add Search(*$Rule_{nr}$*, cons(*Node, SubtrSeq*), $L_j$) to *Agenda*

    • return

4. • create link $L_i = (Node, \{Vertex\})$;
    create $V_i = (Time, State, \{L_i\})$;
    add $L_i$ and $V_i$ to *GSS*

    • for all NewHypo($H$) $\in OldHypoActions$, where $H = (ST, ET, K)$, $ST = Time$,
    for all categories $C_j$ that are possible according to the lexical Key $K$ do

        if $\exists S_j$, st. (Shift $S_j$) $\in$ slr-table(*State, $C_j$*) then

            add Shift($V_i, N_j, ET, S_j$) to *Agenda*, where $N_j = (C_j, ST, ET, .)$

    • $SentinelLink = (\{\}, \{V_i\})$

    • for all categories $C_j$,
    for all (Reduce $Rule_{nr}$) $\in$ slr-table($V_i, C_j$) do

        add Search(*$Rule_{nr}$*, (), *SentinelLink, Time*) to *Agenda*

    • return

NewHypo($H$): $H = (Start, End, Key)$

• for all categories $C_j$ that are possible according to the lexical key *Key* do

    if $\exists N_k = (C_k, S_k, E_k, Hs_k) \in ParseForest$, such that $C_k = C_j$, $S_k = Start$ and $E_k = End$, then

        ∗ add $H$ to $Hs_k$

    else

        ∗ create a node $N_j = (C_j, Start, End, \{H\})$ in the *ParseForest*

        ∗ for all vertices $V_i \in GSS$, such that $V_i = (Time_i, State_i, .)$, $Time_i = Start$, $\exists NewState$ and (Shift $NewState$) $\in$ slr-table($State_i, C_j$), do

            add action Shift($V_i, N_j, End, NewState$) to the *Agenda*

• store NewHypo($H$) in *OldHypoActions*

• return

Search(*$Rule_{nr}$, SubtrSeq, Link, EndingTime*):

1. if $|SubtrSeq| = |\text{right-side}(Rule_{nr})|$ then 2. else 3.

2. if $\exists N_i \in ParseForest$, st. $N_i = (C_i, ST_i, ET_i, StS_i)$, HeadCat($Rule_{nr}$) $= C_i$, $ST_i = T_k$, $V_k = (T_k, ., .)$, $V_k \in PS_l$, $Link = (., PS_l)$, $ET_i = EndingTime$ then

    • add $SubtrSeq$ to $StS_i$; return

    else

    • create a node $N_i = $ (HeadCat($Rule_{nr}$), $T_k$, *EndingTime*, {*SubtrSeq*}) in the *ParseForest*

    • for all vertices $V_m \in PS$, where $Link = (N, PS)$, $V_m = (., S_m, .)$, $C_i = $ HeadCat($Rule_{nr}$) do

        add Shift($V_m, N_i, EndingTime$,
                slr-table($S_m, C_i$)) to *Agenda*

    • return

3. • for all $V_i \in PS$, st. $Link = (., PS)$, $V_i = (.,.,LS_i)$, and for all $L_j \in LS_i$, where $L_j = (N_j, .)$ do

        add Search(*$Rule_{nr}$*, cons($N_j$, *SubtrSeq*), $L_j$, *EndingTime*) to *Agenda*

    • return

The **main routine** is quite simple:

1. vertex $V_0 = (0, 0, \emptyset)$, $GSS = \{V_0\}$.

2. initialize the *Agenda* with one NewHypo action for each word hypothesis of the lattice

3. **until** there are no more actions on the *Agenda* **do**

    take one action from the *Agenda* and carry it out

$$\log P_{normal}(\overline{w} \mid \text{N-Gram, HMM}) = \left\{ \begin{array}{l} \lambda \cdot \log P_{normal}(\overline{w} \mid \text{N-Gram}) \\ \quad + \\ \log P_{normal}(\overline{w} \mid \text{HMM}) \end{array} \right. \quad (1)$$

$$\log P_{\text{inside}}(L) = \left\{ \begin{array}{ll} \log P_{normal}(\overline{w} \mid \text{N-Gram, HMM}), & \text{if } length(\overline{w}) > 0 \\ 0, & \text{if } length(\overline{w}) = 0 \end{array} \right. \quad (2)$$

$$\log P_{outside}(L) = \max_{l \in \texttt{links}(K)} \texttt{normalize} \left( \begin{array}{l} \texttt{denormalize}(\log P_{outside}(l)) + \\ \texttt{denormalize}(\log P_{inside}(L)) + \\ \lambda \cdot \log P(\texttt{first\_word}(L) \mid \\ \quad \texttt{last\_word}(l), \text{Bigram}) \end{array} \right) \quad (3)$$

Besides, $\varepsilon$-productions need not be handled seperately anymore. Therefore the Common Lisp code of an implementation of the revised GLRP has shrunk about 15% compared with an implementation of the original GLRP.

An implementation of the beam search agenda GLRP is available by FTP from faui80.informatik.uni-erlangen.de, "/pub/lisp/parser/glr-lattice-parser.tar.gz". Or send e-mail.

## 3. WORST CASE BEHAVIOUR OF THE REVISED GLRP

The flexibility of the revised GLRP should not incur heavy costs on the runtime behaviour. The crucial places to look for a decrease in performance compared to Tomita's GLRP are the steps (for-loops and existential conditions) where some instances (vertices, previous search actions, etc.) must be retrieved. However, each of these steps either has an equivalent action in Tomita's GLRP or it can be retrieved trivially by some additional information that must be added to the Graph-Structured Stack (For a thorough description of the new data structure please refer to [9]). Therefore the worst case behaviour of the revised GLRP is of the same order as Tomita's.

## 4. BEAM SEARCH

In order to combine the GLRP with heuristic scores it is necessary to define the scoring function and bring arguments why it was chosen.

### 4.1. A Metric For The Beam Search

At least (see also [12]) the following design criteria should be met for the metric:

1. It should combine a bigram model with the acoustic score of the HMM word models.

2. Evaluations of different actions should be comparable.

3. The complete left context information should be considered.

To ensure the comparability of different actions the probabilities of the bigram model and the word model are normalized. The normalization of the bigram model entails a division by the number of operations that have been applied, while the word model probability is divided by the number of time units the word spans (nevertheless, this is an ad-hoc normalization that must be improved in the long run):

$$\log P_{normal}(\overline{w} \mid Bigram) = \frac{\log P(\overline{w} \mid Bigram)}{\#(BigramOperation)}$$

$$\log P_{normal}(\overline{w} \mid HMM) = \frac{\log P(\overline{w} \mid HMM)}{\#(Frames)}$$

Since both schemes are often not drawn from the same test sample and are seldom really equally important, it is useful to combine them with an adjustment parameter $\lambda$ (eq. (1)). The value of this parameter $\lambda$ can be found by experiments or an optimization procedure.

For the purpose of encorporating the left context probability the evaluations are partitioned in *inside evaluations* and *outside evaluations*.

Furthermore, they are defined on the corresponding links instead of the respective word sequences, because the same word sequence may be a continuation of different left contexts and the respective probabilities under these different conditions may vary. I.e. that different links may have the same or different nodes that cover the same word sequence. The left context of a link can easily be identified by its set of predecessor vertices.

The inside evaluation of a link $L$ with a node covering the word sequence $\overline{w}$ is given by equation (2). The outside evaluation of a link $L$ with predecessor vertex $K$ and a bigram model is defined recursively by eq. (3). If $K$ is the start vertex, then in eq. (3) "$\log P_{outside}(l)$" must be substituted by "0" and the "$\texttt{last\_word}(l)$" by "*BEGIN-MARKER*".

The function denormalize causes an extraction of the acoustic and the n-gram score and undoes the normalization over the respective length. This is necessary, since the normalized scores can not be combined directly. normalize is the corresponding inverse function.

If a link has a node that spans several word sequences a maximization over all sequences must be done, since the best analysis is wanted.

### 4.2. Integrating The Beam Search Into The Revised GLRP

In the implemented version the **Shift** actions are scored according to the outside evaluation of the new link. Of course, also the other two types of action could be scored and worked down according to their scores. But for reasons of simplicity and since most **Search** actions only act along paths with good evaluations — almost these alone are constructed by the **Shift** actions — the **Search** actions and the **NewHypo** actions are handled with a stack that has a higher priority than the **Shift** actions.

The beam search strategy itself consists out of two stages:
1. The algorithm works through the lattice time incrementally. Thereby it evaluates all possible actions during each time frame, but processes only those the score of which is within a beam around the best current score. All other actions are saved onto the "PrunedAgenda".
2. If a parse could not be found during stage 1 the actions that were saved onto the "PrunedAgenda" are processed with a best first search.

### 5. EXPERIMENTS

Tests were carried out with a slr-table of a 1560 rules CFG. Ten word lattices where word hypotheses families were already reduced to single word hypotheses and those single hypotheses numbered between 56 and 202 were parsed with the described beam search strategy. In combination with the acoustic scores from the decoder a bigram model with rather high perplexity ($\approx 52$) was used to guide the search. Under these conditions 80% of the lattices could be handled successfully and with one exception the parsing of the recognized structures took less than 4 seconds.

### 6. RELATED WORK

There have been various systems that employ stochastic control, different strategies and — more rarely — GLR-Parsing techniques. E.g.: Shikano [8] shows how to use n-gram models. Päseler et. al. [3] published a beam search method. This method is combined with a n-gram model by Päseler and Ney [2]. Wrigley and Wright, e.g. [13], use a probabilistic CFG as their language model. In [6] best first search is demonstrated by T. Seneff. In H. Ney [1] sets of phoneme hypotheses are analysed with a beam search strategy and the use of n-gram models in this context is explained. L. Schmid uses the A*-algorithm to do a best first search over the number of word hypotheses in [5]. Etc.

However, none of them presents *a general approach* to guide the GLR-parsing process with stochastic information and especially to combine GLR-parsing with a beam search strategy.

### 7. CONCLUSION

In this paper a revised version of Tomita's GLR-parsing algorithm is described that allows the flexible use of strategies. It is combined with a beam search strategy to parse word lattices and return a packed forest representation of a number of parse trees with good scores. While theoretical considerations show that the worst case behaviour of the revised algorithm is of the same order as Tomita's original algorithm the experiments demonstrate that the new algorithm might be used in nontrivial applications.

The experiments also indicate that GLRP itself is useful. E.g. Schabes [4] argues that GLRP was heavily handicapped, because the number of slr-table states could be exponential to the number of CFG rules. However, the 1560 rules CFG that is used in the experiments generates only ca. 4500 states. This supports Tomita's claim, that grammars for natural languages were not too densely ambiguous and therefore GLRP was appropriate for natural language parsing.

### Acknowledgements

This research is part of the author's Master's thesis. It was done at the department of computer science VIII (AI), University of Erlangen, and encouraged by my advisor Mark Steedman, University of Pennsylvania. I would like to thank Hans Weber for many fruitful discussions and Günther Görz and Angela Rösch for helpful comments on the paper.